\documentclass{article}

\usepackage[preprint, nonatbib]{nips_2018}

\usepackage{float}

\usepackage[square,numbers]{natbib}
\usepackage[utf8]{inputenc} 
\usepackage[T1]{fontenc}    
\usepackage{hyperref}       
\usepackage{url}            
\usepackage{booktabs}       
\usepackage{amsfonts}       
\usepackage{nicefrac}       
\usepackage{microtype}      
\usepackage{graphicx}
\usepackage{tabularx}
\usepackage{natbib}
\usepackage{hyperref}
\usepackage{amsmath}

\title{Multi-Step Embed to Control: A Novel Deep Learning-based Approach for Surrogate Modelling in Reservoir Simulation}

\author{
  Jungang Chen, Eduardo Gildin, John Killough\\
  Harold Vance Department of Petroleum Engineering, Texas A\&M University, USA\\
  \texttt{\{jungangc, egildin, jkillough\}@tamu.edu} \\
  }

\begin{document}
\maketitle

\begin{abstract}
	Reduced-order models, also known as proxy model or surrogate model, are approximate models that are less computational expensive as opposed to fully descriptive models. With the integration of machine learning, these models have garnered increasing research interests recently. However, many existing reduced-order modeling methods, such as embed to control (E2C) and embed to control and observe (E2CO), fall short in long-term predictions due to the accumulation of prediction errors over time. This issue arises partly from the one-step prediction framework inherent in E2C and E2CO architectures. This paper introduces a deep learning-based surrogate model, referred as multi-step embed-to-control model, for the construction of proxy models with improved long-term prediction performance. Unlike E2C and E2CO, the proposed network considers multiple forward transitions in the latent space at a time using Koopman operator, allowing the model to incorporate a sequence of state snapshots during training phrases. Additionally, the loss function of this novel approach has been redesigned to accommodate these multiple transitions and to respect the underlying physical principles. To validate the efficacy of the proposed method, the developed framework was implemented within two-phase (oil and water) reservoir model under a waterflooding scheme. Comparative analysis demonstrate that the proposed model significantly outperforms the conventional E2C model in long-term simulation scenarios. Notably, there was a substantial reduction in temporal errors in the prediction of saturation profiles and a decent improvement in pressure forecasting accuracy.
\end{abstract}


\section{Introduction}
\label{sec:introduction}
Reservoir simulation is a computational process used to model multi-physics behaviors of fluids within the subsurface reservoir over a specified spatiotemporal domain. It primarily focuses on utilizing mathematical models to predict the flow of oil, gas and water through porous media. Recent efforts have expanded to include thermos-hydro-mechanical interactions and applications in geothermal energy, CO2 sequestration and hydrogen storage. However, full scale simulations, especially those modeling multi-physics, are computationally intensive and time-consuming, making them impractical for real-time decision-making, optimization, uncertainty quantification, and frequent data assimilation.

To address these challenges, reduced-order models have been introduced to simplify those complex simulations, enabling faster and more efficient decision making while maintaining essential accuracy. Traditional reduced-order modeling (ROM) techniques, such as those using Proper Orthogonal Decomposition (POD), seek to project the original high-dimensional space $x$ to a lower-dimensional latent state space $z$ using the POD basis $\Phi$ derived from collected data snapshots, e.g. $z = \Phi x$ \cite{trehan2016trajectory, tan2019trajectory}. Despite the effectiveness, their reliance on direct source code access could hinder their practical applications. Alternatively, deep learning-based ROMs offer a non-intrusive approach that can seamlessly integrate with existing commercial simulators \cite{jin2020deep, coutinho2021physics}. Moreover, traditional ROMs assume that the dynamics of the system can be accurately captured in a linear subspace of the full state space. This assumption can limit its effectiveness in handling systems with highly nonlinear behavior. On the other hand, deep learning-based ROMs leverage neural networks to account for nonlinear dynamics.

With the development of machine learning techniques and their application to subsurface problems \cite{li2023rapid,yan2023estimation, zhang2023application,li2018contextual}, various machine learning-based surrogate models have been proposed for different settings \cite{tang2022deep, zheng2024deep, chen2023physics}. However, the embed to control (E2C) model and its variants are particularly suitable for scenarios with time-varying well settings, such as bottom-hole pressure and injection rates, where well-control optimization is involved. The original Embed-to-Control (E2C) is a variational autoencoder-based deep learning model for stochastic optimal control \cite{watter2015embed}. It has been modified to a deterministic form for surrogate modeling in reservoir simulation \cite{jin2020deep, coutinho2021physics}. This model consists of an encoder network that projects high-dimensional state spaces (saturation, pressure, etc.) into a lower latent space, a transition model that governs the evolution of this latent space from current time to the next, and a decoder that reconstructs states of next time from latent representation of corresponding time step. However, the conventional E2C model, primarily designed for short-term forecasting due to its reliance on adjacent state pairs for training, encounters limitations in precise long-term simulation scenarios.

In this paper, we propose a novel deep learning-based ROMs that is more accurate in terms of long-term predictions. Similar to the variational autoencoder structure in E2C, the newly proposed model also employs an encoder that projects original high-dimensional state to a lower-dimensional latent representation and a decoder that reconstruct the state using predicted latent state. However, unlike the locally linear transition model in E2C, we use Koopman operator to perform multiple forward transitions in latent space simultaneously, which allows the network to make multi-step state predictions by decoding predicted latent representations. Notably, the Koopman operator is also approximated by neural networks, so the overall network is trained in an unsupervised manner. Furthermore, the loss function has been modified to account for the multi-step prediction loss. We note these modifications improve the long-term prediction accuracy. This enhancement can be primarily attributed to the integration of multiple transitions within the model's framework, which effectively mitigates the accumulation of error along the trajectory. By improving the accuracy of long-term predictions of states and observations, the model can facilitate the development of more reliable well rate/pressure control strategies for longer reservoir production period \cite{chen2024optimization}.

This paper is organized as follows: we begin with a review of deep generative models and show their relationship with reduced-order modeling. We then highlight the limitations of previous models in long-term simulations and introduce our model which is more accurate for long-term predictions. Following that, we present and compare our results with previous models. Finally, we conclude this work and discuss future research directions.

\section{Problem Statement}
\label{sec:problem-statement}

Instead of comparing deep-learning ROMs with traditional ROM techniques, in this section, we present the deep learning-based ROMs within the framework of deep generative models for sequential data. The rational behind is that given a sequence of control inputs and the initial state, what would be the most probable resulting state trajectory? Mathematically, it can be formulated as $p(x_{1:T}|u_{1:T}, x_o)$. Once $p(x_{1:T}|u_{1:T}, x_o)$ is obtained through deep learning, the trained model can generate the most probable trajectory $x_{1:T}$ for any given testing sequence of control inputs $u_{1:T}$ and initial state $x_0$. Although the generation process is stochastic, it can be easily converted to deterministic form, as in reservoir simulation case, by modifying the encoding and decoding network architectures. The model is called 'reduced order' because it leverages the encoder-decoder structure to compress the original state into a latent space and reconstruct original states using predicted latent states.

\subsection{Variational autoencoder (VAE)}
Variational autoencoder is a type of deep generative model that employs an encoder-decoder structure to generate new data samples similar to input data it has been trained on \cite{kingma2013auto, chen2023generating, misra2023massive, ding2024enhance}. VAE-based generative models often consists of two processes: generative process and inference process. The generative process assumes that the observed data x is generated from latent variable z which has a simpler distribution (e.g. standard Gaussian). Once z has been sampled from the prior distribution $p(z)$, the new data can be generated from the conditional probability $p_\theta(x|z)$ which is parameterized by a neural network with parameter $\theta$. The inference process involves calculation of the posterior distribution $p_\theta(z|x) = \frac{p_\theta(x|z)p(z)} {p(x)} $. However, directly computing the posterior distribution is intractable because it requires integration over latent variable z. To tackle this issue, VAEs adopts variational inference to approximate the posterior distribution using a distribution which is parameterized by neural network, expressed as $q_\varphi(z|x)$.

To approximate the true posterior, the VAE is trained to maximize the evidence lower bound (ELBO), which reads,
\begin{equation}
    \mathcal{L}(\theta, \varphi, x)
    = \mathbb{E}_{q_\varphi(z|x)}[\log(p_\theta(x|z))]-D_{KL}(q_\varphi(z|x))\parallel p(z)),
    \label{eq:elbo_vae}
\end{equation}
The first term is referred as reconstruction, which measures how well the data can be reconstructed through the encoder-decoder process. The second term is called regularization term, which enforces the approximate posterior $q_\varphi(z|x)$ to be close to the prior $p(z)$.

Despite their effectiveness, VAEs are not designed for sequential data \cite{girin2020dynamical}. In fact, VAE assumes the input data being independent and identically distributed (i.i.d.), which hinders its application to sequential data. Furthermore, VAEs cannot handle system controls, therefore it is not suitable for dynamical systems where both system inputs and state transitions need to be considered.

\subsection{Embed to Control (E2C)}
The original Embed-to-Control (E2C) is a variational autoencoder-based deep learning model for stochastic optimal control of dynamical systems \cite{watter2015embed}. Similar to VAE, the E2C has an encoder-decoder structure. Instead of modeling data likelihood $p(x)$, the E2C models the pair likelihood $p(x_t, x_{t+1} | u_t)$, where $x_{t+1}$ is a result from $x_t$ by applying $u_t$. E2C assumes $x_t$ and $x_{t+1}$ can be generated from latent variable $z_t$ and $z_{t+1}$ respectively. To consider state transitions, it enforces $z_{t+1}$ to be both the embedding of $x_{t+1}$ and the evolution of $z_t$ by applying $u_t$. The latent dynamics in general is expressed as:

\begin{equation}
    \hat{z}_{t+1} = f_z(z_t, u_t),
    \label{eq:transition}
\end{equation}
Where $f_z$ signifies the latent dynamics operator that evolves the latent state based on current state $z_t$ and control input $u_t$. Accordingly, the E2C aims to maximize the following evidence lower bound (ELBO) \cite{banijamali2018robust}:

\begin{equation}
    \begin{aligned}
        \mathcal{L}(\theta, \varphi, \omega, x_t, x_{t+1})
        =\mathbb{E}_{q_\varphi(z_t|x_t), q_\varphi(z_{t+1}|x_{t+1})}[\log(p_\theta(x_t|z_t))+\log(p_\theta(x_{t+1}|z_{t+1}))] \\
         -D_{KL}(q_\varphi(z_t|x_t))\parallel p(z_t)) -D_{KL}(q_\omega(\hat{z}_{t+1}|z_t,u_t))\parallel q_\varphi(z_{t+1}|x_{t+1})),
     \label{eq:elbo_e2c}
    \end{aligned}
\end{equation}

Where $q_\omega(\hat{z}_{t+1}|z_t,u_t)$ represents the generative dynamics of equation \ref{eq:elbo_e2c} which is parameterized by neural network with weights $\omega$. In practice, it’s usually replaced by a locally linear formulation. Likewise, the first term in equation \ref{eq:elbo_e2c} represents the reconstruction term which measures how well $x_t$ and $x_{t+1}$ can be reconstructed, the second and last term means that not only it enforces the posterior $q_\varphi(z_t|x_t)$ to be like the prior $p(z_t)$, it also enforces the result $\hat{z}_{t+1}$ from the latent dynamics to be close to the embedding $z_{t+1}$.

\cite{jin2020deep} has modified the above formulation to a deterministic version for surrogate modelling of reservoir simulation applications. This model consists of an encoder network that projects high-dimensional states into a lower latent space, a transition model that governs the evolution of this latent space from current time to the next, and a decoder that reconstructs states of next time from latent representation of corresponding time step. From a mathematical standpoint, the encoder-transition-decoder architecture, is detailed as follows:




    \makebox[5.5cm]{Encoder:}  $z_t = Q_\varphi(x_t)$\par
    \makebox[5.5cm]{Transition: }  $\hat{z}_{t+1} = F_{\omega_1}(z_t, u_t)$\par
\begin{figure}[H]

    \centering
    \includegraphics[width=0.5\linewidth]{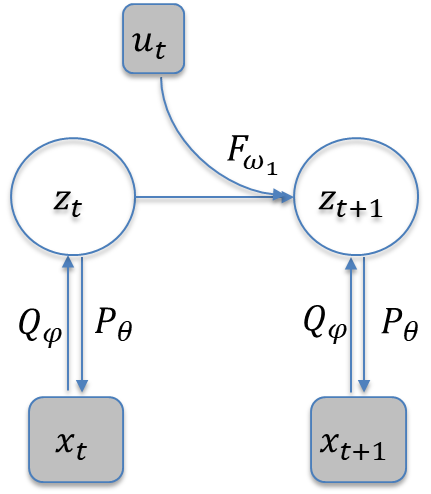}

    \caption{Graphical model of E2C.}
    \label{fig:e2c}
\end{figure}

    \makebox[5.5cm]{Decoder: }  $\hat{x}_{t+1} = P_\theta(\hat{z}_{t+1})$\par
where $Q_\varphi$ and $P_\theta$ denote the deterministic form of $q_\varphi$and $p_\theta$ , and $F_{\omega_1}$ represents the deterministic latent state dynamics. A graphical representation of E2C is presented in Figure \ref{fig:e2c}.

In practice, the latent dynamics is usually replaced by a locally linear form:

\begin{equation}
    \hat{z}_{t+1} = A(z_t)*z_t+B(z_t)*u_t,
    \label{eq:locally_linear}
\end{equation}

Although effective, E2C is not designed for long trajectory prediction, as indicated by equation \ref{eq:elbo_vae}, E2C aims to maximize the pair marginal likelihood. In other words, for a given $x_t$ and $z_t$, E2C tries to predict the most probable $x_{t+1}$, which referred as one-step prediction. For long-term trajectory prediction, the one-step prediction will naturally lead to error accumulation problem.

\section{Methods}
\label{sec:methods}
Unlike E2C, the newly proposed model performs multi-step forward prediction through a deep Koopman operator. Instead of relying on single-step transition in the latent space and utilizing adjacent states for training, the new framework employs a multiple transition strategy. This approach incorporates multiple state snapshots and control data to facilitate the learning of a more refined latent space, thereby significantly improving the accuracy of long-term predictions. The framework of the newly proposed models is presented in Figure \ref{fig:multistep_e2c} below:

\begin{figure}[H]
    \centering
    \includegraphics[width=\linewidth]{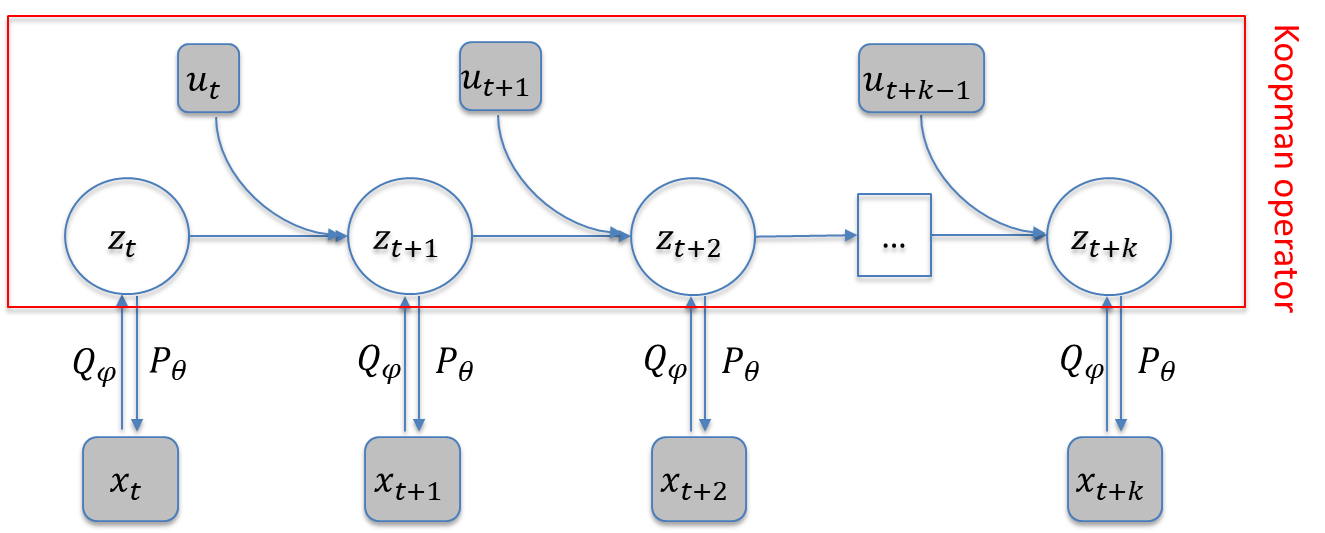}

    \caption{Framework of proposed multi-step E2C model. The latent space is further lifted by Koopman operator to a space where the state transition is linear. The network can be equipped with multiple state snapshots for training and therefore improving the long-term prediction accuracy.}
    \label{fig:multistep_e2c}
\end{figure}

In this proposed approach, it is assumed that a sequence of states $x_{t}$, $x_{t+1}$, ..., $x_{t+k}$ can be generated by corresponding latent states $z_{t}$, $z_{t+1}$, ..., $z_{t+k}$ with the decoder structure. The key distinction of our method lies in how the latent space is propagated. \cite{tytarenko2022multi} attempted to use the E2C approach but found that special treatment was needed to maintain stability. Unlike the locally linear form used in E2C, where locally A and B matrices are applied to transitions (by locally we meant that A and B matrices in \ref{eq:locally_linear} change with different latent states), our model leverages the Koopman operator to project the latent space into a higher-dimensional space where multi-step transitions become globally linear, as in \ref{eq:koopman_multi_transition}. As a result, our model, referred to multi-step embed to control (MSE2C), models the likelihood $p(x_t, x_{t+1}, ...x_{t+k} | u_t, u_{t+1}, ...u_{t+k-1})$, where $k$ is the forward step. The resulting loss of the whole framework $\mathcal{L}_{total}$ is written as follows(the details are presented in next section): 

\begin{equation}
    \begin{aligned}
        \mathcal{L}_{total}
        =\mathcal{L}_{rec}+\mathcal{L}_{reg}+\mathcal{L}_{koop}
     \label{eq:koopman_loss}
    \end{aligned}
\end{equation}

\subsection{Deep Koopman Operator}
The Koopman operator is a powerful tool to analyze nonlinear dynamical systems by mapping them into an infinite-dimensional framework \cite{koopman1931hamiltonian}. The basic concept of Koopman operator is to shift the focus from the state space to the space of observable functions where linear operator can be applied. Practical implementations, such as dynamic mode decomposition (DMD) and its variant extended DMD, seek to approximate the Koopman operator using data from the system. For more details regarding Koopman operators, we refer to \cite{proctor2018generalizing, williams2015data, schmid2010dynamic}. Recent advancements involve using deep learning to automatically learn the appropriate observables and Koopman operator representations from data \cite{morton2019deep, shi2022deep}(Morton, J et al., 2019; Shi, H et al., 2022). \cite{xiao2022deep} proposed a data-driven approach to model and control autonomous vehicle with Koopman operators approximated by neural network. However, directly applying Koopman operators to original states may not be effective for high-dimensional systems. In this paper, instead of applying Koopman operators directly to the original states, we lift the latent space to observables using neural networks, which we believe is more effective for high-dimensional systems. A draft structure is presented in Figure 3.
\begin{figure}[H]
    \centering
    \includegraphics[width=\linewidth]{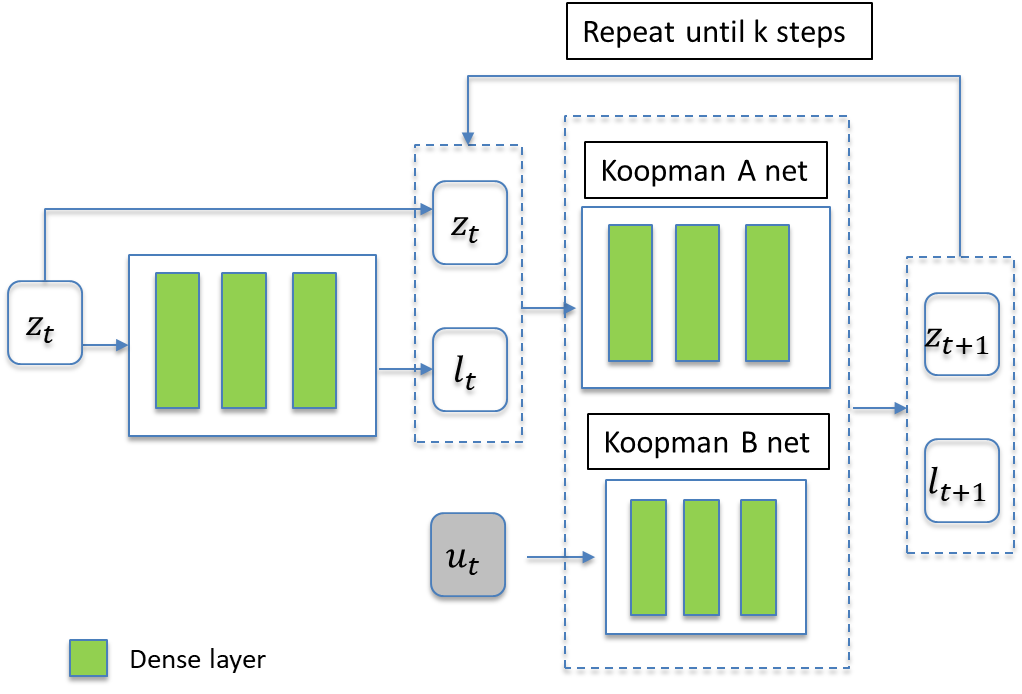}

    \caption{The structure of deep Koopman operator, where green box represents dense layers.}
    \label{fig:deepkoopman}
\end{figure}

Again, let $\hat{z}_{t+1} = f_z(z_t, u_t)$ be the nonlinear transition dynamics of the E2C. The deep Koopman operator tried to lift the latent space representation $z_t$ to a space where the following operation is linear:

\begin{equation}
    \Phi_{e,t+1}=A\Phi_{e,t}+Bu_t
    \label{eq:koopman_transition}
\end{equation}

Where $\Phi_{e,t}=[z_t, l_t]^{T}$, and $l_t$ denotes the output of last layer of the encoder in Figure \ref{fig:deepkoopman}, which reads as $l_t=enc(z_t)$. The matrix $A$ and $B$ (we slightly abuse the notation here and note that these are different with what in the locally linear form), are defined as the approximate Koopman operator with finite dimension. For brevity, equation \ref{eq:koopman_transition} can be rearranged as $\Phi_{e,t+1}=K\Psi_{t}$. $K=[A, B]$ is the Koopman operator approximated by neural network weights, and $\Psi_{t}=[\Phi_{e,t}, u_t]^{T}$.

With approximate Koopman operator, multi-step forward prediction can be performed in the lifted space, which follows:

\begin{equation}
    \Phi_{e,t+p}=A\Phi_{e,t+p-1}+Bu_{t+p-1} \qquad (p = 1, 2, \dots, k)
    \label{eq:koopman_multi_transition}
\end{equation}

To learn the Koopman operator, the error of the state evolution in the lifted space and those from observations needs to be minimized. To this end, the following k-step loss is minimized:
\begin{equation}
    \begin{aligned}
        \mathcal{L}_{koop}
        =\frac{1}{k}\sum_{i=1}^{k}||\hat{\Phi}_{e,t+i}-K^{[i]}\Psi_{t}||^{2},
     \label{eq:koopman_loss}
    \end{aligned}
\end{equation}

Where $\hat{\Phi}_{e,t+i}=[Q_\varphi(x_{t+i}), enc(Q_\varphi(x_{t+i}))]^{T}$ are mapped from observations, and $K^{[i]}\Psi_{t}$ are predicted from the Koopman operator.

\subsection{Multi-step E2C training}
In this section, we outline the loss functions used to train the multi-step E2C network. The loss consists of 3 components: reconstruction loss that enforces the predicted states close to the true states, regularization loss to ensure the encoded latent states are also results of k-step prediction from Koopman operator, and the k-step loss that uses to approximate the Koopman operator. The first two losses are similar to the VAE loss in equation \ref{eq:elbo_vae} and E2C loss in equation \ref{eq:elbo_e2c}, but note that in this context it’s different sign as we minimize the loss instead of maximizing likelihood. The k-step loss follows the definition in equation \ref{eq:koopman_loss}. In summary, the loss terms are defined as:

        

        

\begin{equation}
    \begin{aligned}
        \mathcal{L}_{rec}
        =-\sum_{i=1}^{k}\mathbb{E}_{q_\varphi(z_{t+1}|x_{t+1})}[\log(p_\theta(x_{t+1}|z_{t=1}))],
     \label{eq:rec_loss}
    \end{aligned}
\end{equation}

\begin{equation}
    \begin{aligned}
        \mathcal{L}_{reg}
        =D_{KL}(q_\varphi(z|x))\parallel p(z))
        + \sum_{i=1}^{k}D_{KL}(k_\omega(\hat{z}_{t+i} | z_t, u_t ) \parallel q_\varphi(z_{t+i}|x_{t+i}))
     \label{eq:reg_loss}
    \end{aligned}
\end{equation}

\begin{equation}
    \begin{aligned}
        \mathcal{L}_{koop}
        =\frac{1}{k}\sum_{i=1}^{k}||\hat{\Phi}_{e,t+i}-K^{[i]}\Psi_{t}||^{2},
     \label{eq:koopman_loss2}
    \end{aligned}
\end{equation}

where $k_\omega(\hat{z}_{t+i} | z_t, u_t )$ in $\mathcal{L}_{reg}$ denotes the latent state predictions from Koopman operator.To sum up, we present the training tuple and training loss of different methods in Table \ref{tab:comparison}. It can be observed that one obvious advantage of our approach is that more data considered for training, which significantly reduces the error accumulation along the simulation trajectory.

The training process consists of 400 epochs with a batch size of 6. The Adam optimizer is used for training, with a learning rate set at 0.01. The data is split into training and testing sets with a ratio of 4:1.

\begin{table}
  \caption{Comparison of training tuple and training loss for different methods}
  \label{sample-table}
  \centering
  \begin{tabular}{llc}
    \toprule
    Methods    & Training tuple    & Training losses\footnotemark  \\
    \midrule
    VAE &   \{$x$\}      & \{$\mathcal{L}_{rec}$,$\mathcal{L}_{reg}$\}   \\
    Embed to Control (E2C)  & \{$x_t$,$u_t$,$x_{t+1}$\}         & \{$\mathcal{L}_{rec}$, $\mathcal{L}_{reg}$\}   \\
    Multi-step E2C  & \{$x_t$,$u_t$,$x_{t+1}$,$u_{t+1}$,...,$x_{t+k}$\}       & \{$\mathcal{L}_{rec}$, $\mathcal{L}_{reg}$, $\mathcal{L}_{koop}$ \}  \\
    \bottomrule
  \end{tabular}
  \label{tab:comparison}
\end{table}
\footnotetext{We abuse the loss term here.$\mathcal{L}_{rec}$ and $\mathcal{L}_{reg}$ represent the reconstruction and regularization loss of each unique approach.}

\section{Results}
\label{sec:results}
\begin{figure}[H]
    \centering
    \includegraphics[width=0.8\linewidth]{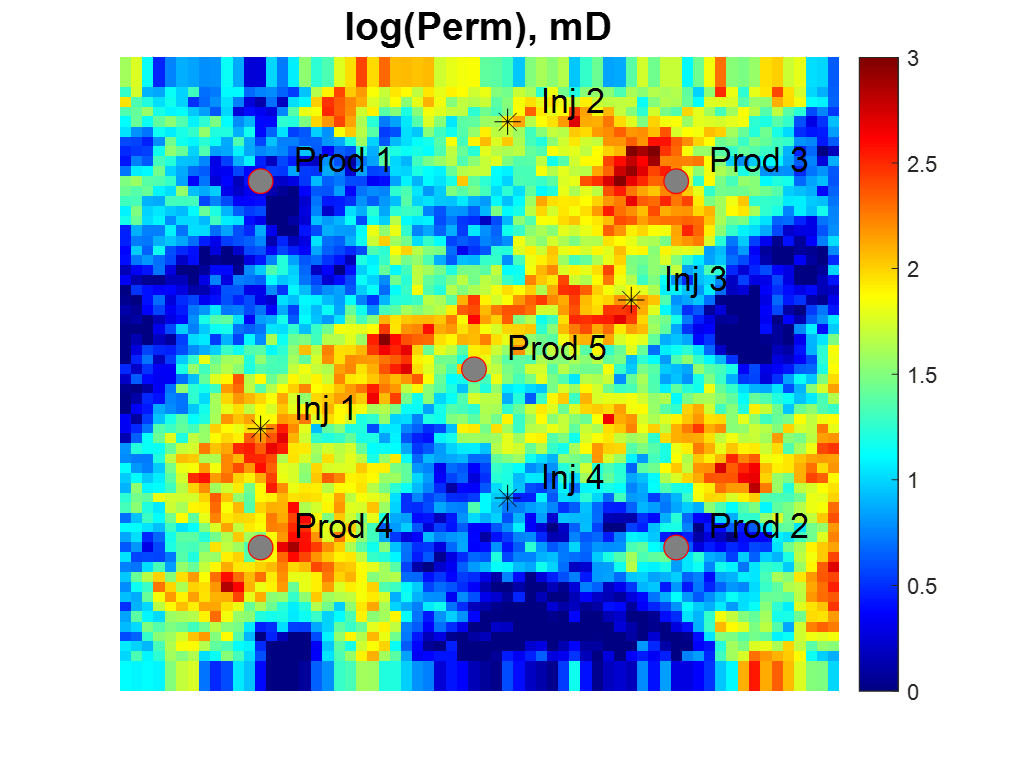}

    \caption{Permeability map and well locations.}
    \label{fig:perm_map}
\end{figure}

The proposed framework is applied within a water flooding case where water is injected into an oil reservoir. The model consists of 64 × 64 × 1 grid blocks, each of which is of the same size and has dimension of 65.6 ft × 65.6 ft × 65.6 ft in x, y and z directions. We simulated scenarios involving the water injection using 4 injector wells and the extraction of oil using 5 producer wells. The well locations and permeability map are plotted in Figure \ref{fig:perm_map}. This model runs for 2000 days with well controls, including water injection rates and bottom hole pressures (BHPs), changing every 100 days.

In this experiment, a total of 600 simulation runs are performed during the offline stage to gather data. Each simulation run, also referred as one trajectory, consists of 21 state (pressure, saturation) snapshots and 20 well controls (BHP, injection rate). For a k-step model, we can collect $21-k+1$ sets of data per trajectory, resulting in a total of $600*(21-k+1)$ data samples for all simulation runs.

Three approaches, including the E2C, multistep E2C with $k=5$, and multistep E2C with $k=7$, are compared to validate the efficacy of proposed approach.

\subsection{Comparison of E2C and Multi-step E2C}
\begin{figure}[H]
    \centering
    \includegraphics[width=\linewidth]{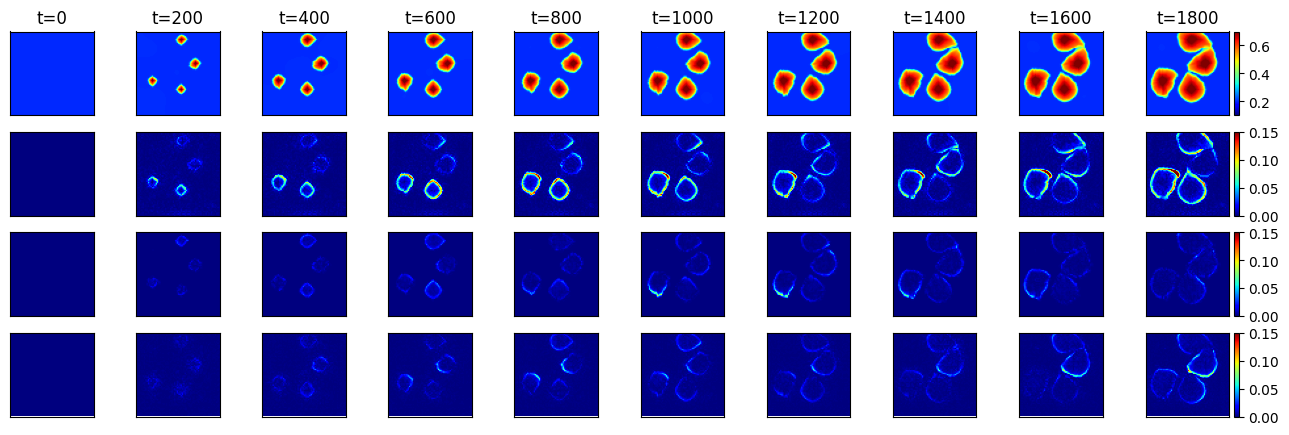}

    \caption{Comparison of E2C, multi-step E2C (k=5) and multi-step E2C (k=7) predictions on water saturation evolution at different days.}
    \label{fig:sat_map}
\end{figure}

\begin{figure}[H]
    \centering
    \includegraphics[width=\linewidth]{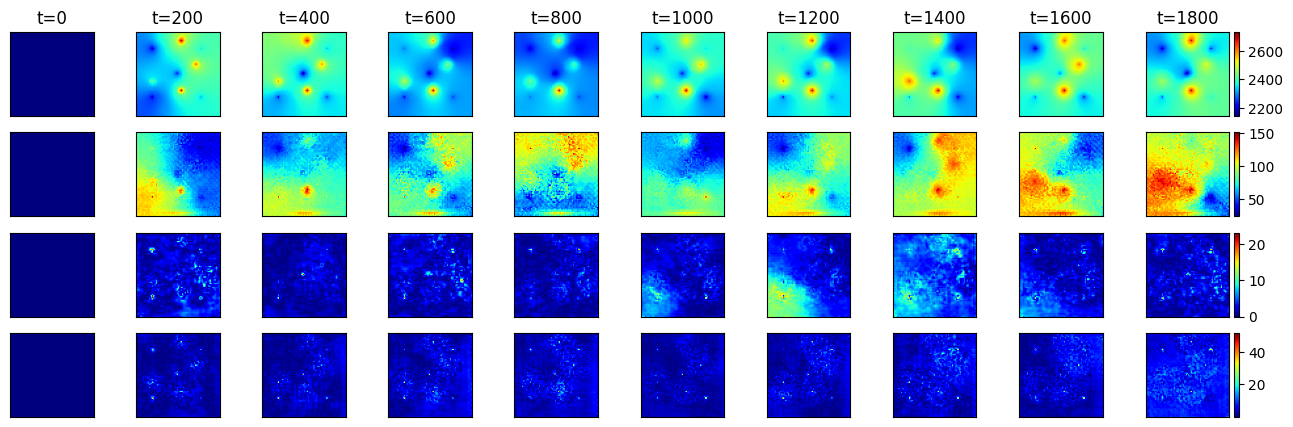}

    \caption{Comparison of E2C, multi-step E2C (k=5) and multi-step E2C (k=7) predictions on fluid pressure evolution at different days.}
    \label{fig:pres_map}
\end{figure}

Figure \ref{fig:sat_map} and \ref{fig:pres_map} compares the water saturation and fluid pressure predicted by E2C, multi-step E2C with a forward prediction step of 5 (k=5) and multi-step E2C with a forward prediction step of 7 (k=7), against those obtained from numerical simulations.

The top row of Figure \ref{fig:sat_map} shows the water saturation at different times from numerical simulation. The second row of Figure \ref{fig:sat_map} depicts the absolute error of water saturations between the numerical simulation and those predicted by E2C at corresponding timesteps. The third row and last row present the absolute error of water saturation predictions between the numerical simulation and multi-step E2C (k=5) and multistep E2C (k=7) respectively. It can be observed that both multi-step E2C methods perform better than the vanilla E2C at every time steps.

Similarly, we present the fluid pressure and its absolute error map from different approaches at difference time in Figure \ref{fig:pres_map}. The top row represents the fluid pressures from numerical simulator, while the second row presents the corresponding absolute error map predicted by E2C. The last two rows depict the absolute error map predicted by multistep E2C (k=5) and multistep E2C (k=7) respectively. It can be also observed that the multi-step E2C achieves significantly better long-term accuracy for predicting fluid pressures.

\begin{figure}[H]
    \centering
    \includegraphics[width=\linewidth]{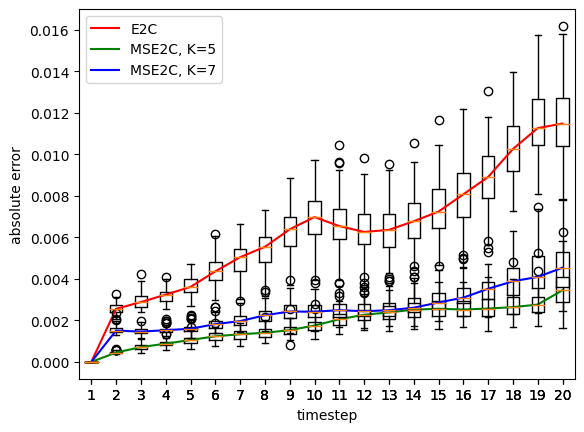}

    \caption{Boxplot of absolute error along with timesteps for different approaches.}
    \label{fig:boxplot_map}
\end{figure}

\begin{figure}[H]
    \centering
    \includegraphics[width=\linewidth]{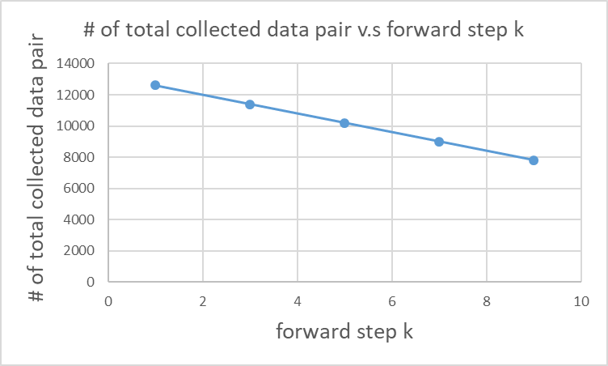}

    \caption{Collected samples decreases as forward step $k$ increases.}
    \label{fig:sample_vs_kstep}
\end{figure}

Finally, to demonstrate the effectiveness of the proposed multi-step E2C across various testing scenarios, Figure \ref{fig:boxplot_map} presents a boxplot that shows the absolute error of 100 testing cases over time for different approaches. As the plot illustrates, the E2C approach consistently exhibits the highest absolute error across timesteps, indicating it is less accurate than the other methods. In contrast, the multi-step E2C approaches with k=5 and k=7 show similar accuracy, with the multi-step E2C (k=5) performing slightly better. This slight performance difference can be attributed to the reduction in training samples when the forward timesteps are increased(as presented in \ref{fig:sample_vs_kstep}), making the network more challenging to train effectively.

\section{Conclusions}
\label{sec:conclusion}
This study introduces a novel deep learning-based reduced-order model designed for efficient long-term predictions in reservoir simulations. The proposed multi-step E2C method incorporates multiple forward transitions in the latent space using the Koopman operator, significantly enhancing the accuracy of long-term predictions by mitigating error accumulation typically observed in conventional single-step E2C models. The efficacy of the proposed method is validated using a two-phase (oil and water) reservoir model under a water flooding scheme. Comparative analysis highlights that the multi-step E2C model outperforms conventional E2C in both water saturation and fluid pressure predictions over extended simulation periods. By improving the accuracy of long-term predictions, the proposed model can facilitate more reliable well rate and pressure control strategies, ultimately contributing to more effective reservoir management and decision-making processes. In conclusion, the multi-step E2C model offers a robust and computationally efficient solution for long-term reservoir simulation and optimization, addressing key limitations of existing reduced-order modeling techniques.

\subsubsection*{Acknowledgments}
The authors would like to thank Texas A\&M High Performance Research Computing for advanced computing resources.

\bibliography{references.bib}

\end{document}